# Power Transformer Fault Diagnosis with Intrinsic Time-scale Decomposition and XGBoost Classifier

Shoaib Meraj Sami[1] and Mohammed Imamul Hassan Bhuiyan[2]

[1,2]Department of Electrical and Electronic Engineering,
Bangladesh University of Engineering and Technology, Dhaka 1205, Bangladesh
[1]shoaib.eee08@gmail.com
[2]imamul@eee.buet.ac.bd

**Abstract.** An intrinsic time-scale decomposition (ITD)-based method for power transformer fault diagnosis is proposed. Dissolved gas analysis (DGA)parameters are ranked according to their skewness, and then ITD based features extraction is performed. An optimal set of PRC features are determined by an XGBoost classifier. For classification purpose, an XGBoost classifier is used to the optimal PRC features set. The proposed method's performance in classification is studied using publicly available DGA data of 376 power transformers and employing an XGBoost classifier. TheProposed method achieves more than 95% accuracy and high sensitivity and F1-score, better than conventional methods and some recent machine learning-based fault diagnosis approaches. Moreover, it gives better Cohen Kappa and F1-score as compared to the recently introduced EMD-based hierarchical technique for fault diagnosis in power transformers.

## 1    Introduction

Power transformer is one of the most essential equipments for power transmission and distribution system. Monitoring the condition and fault diagnosis is very important for ensuring uninterrupted electricity supply. Due to thermal, electrical stress of insulation material and aging a variety of faults occur in power transformers. These faults have strong correlation with concentration of the dissolved gas emitted from the oil or cellulose paper. The gases include hydrogen ($H_2$), methane ($CH_4$), ethane ($C_2H_6$), ethylene ($C_2H_4$) and acetylene ($C_2H_2$). Dissolved gas analysis (DGA) methods such as Duval Triangle, Rogers Ratio, IEC method and Dornenburg ratio method are widely used for the detection of power transformer faults [1-3]. A limitation of these methods is occasional poor performance and ambiguity to detect fault. To overcome these shortcomings, many machine learning-based approaches are proposed in the literature [4-7].

Recently, empirical mode decomposition (EMD) based feature extraction from ranked features is shown to provide promising results for transformer fault detection[1] [5]. However, compared to EMD, intrinsic time-scale decomposition (ITD) has several benefits because ITD consider proper rotational property of any non-stationary signal [7]. As such it can provide more information while being computationally efficient and more robust to noise [9-10]. Interestingly, ITD has also been successfully to analyze a variety of non-stationary signals and related prediction with machine-learning [10-13].

To the best of our knowledge, the use of ITD for the detection of transformer faults using machine-learning is yet to be reported. In this paper, a number of DGA parameters used in the traditional methods are first ranked by their skewness. A subset of the DGA parameters in terms of their increasing rank are decomposed into the ITD domain. The resulting PRC components are used as features and classified by an XGBoost classifier for transformer fault detection. Note that unlike [5], the proposed approach employs a single XGBoost classifier thus, reducing complexity in the classification system and computational time. The performance of the proposed Method is studied using publicly available DGA data of 376 transformers and compared with those of recently reported results.

## 2    The DGA Dataset

The DGA data of 376 power transformers is used. Among them, data of 239 transformers are collected from publicly available Egyptian Electricity Network samples [14]. Others are collected from different published scientific literatures [2,14]. Six different types of faults are used. These are partial discharge (PD), low energy discharge (D1), high energy discharge (D2), and three different types of thermal faults those are T1 (temperature less than 300°C), T2 (temperature between 300°C to 700°C) and T3 (temperature greater less than 700°C). A summary of the DGA data is provided in Table-I.

Table I: Distribution of different fault types

| Fault Type | PD | D1 | D2 | T1 | T2 | T3 | Overall |
|---|---|---|---|---|---|---|---|
| Lab Samples | 27 | 42 | 54 | 70 | 18 | 28 | 239 |
| Literatures Sample | 15 | 25 | 59 | 10 | 3 | 25 | 137 |
| Total | 42 | 67 | 113 | 80 | 21 | 53 | 376 |

Notice that  the dataset  is slightly unbalanced because about 30% of samples belongs to  D2 fault and 5.59%   to  T2 fault.  This will be taken into account during performance analysis of the Proposed Method described in Section 4.
Methodology
In this Section DGA parameter generation and their ranking, ITD-based feature extraction and optimal feature selection are described. Hydrogen ($H_2$), methane ($CH_4$), ethane ($H_2$), ethylene ($CH_4$) and acetylene ($C_2H_2$) are main DGA gas parameter for this purpose.

### 2.1    Ratio-based DGA Parameter Generation and Ranking

In the literature, many ratio-based DGA parameters are generated from Hydrogen ($H_2$), methane ($CH_4$), ethane (), ethylene ($C_2H_4$) and acetylene ($C_2H_2$). We used thirty-seven DGA parameters from this work, which is collected from literature [5]. Those parameters are illustrated in Table-II. Different ratio-based parameters provide different discriminatory properties of faulty transformer.

After generating different DGA ratio-based parameters, we rank them by skewness. For this purpose, we calculate skewness of each parameter for the 376 transformers. After thatthe parameters are ranked from lower skewness to higher skewness. Similar work is also performed in the literature [5]. This ranking scenario is presented in Table-II.We also investigate the distribution of transformer DGA parameter among six classes of faults using Box plots (Fig.1). One can see that with increasing skewness the distribution of DGA parameters become more indistinguishable and ambiguous.

Table II: The DGA Parameters

| No. | Parameter | No. | Parameter | No. | |
|---|---|---|---|---|---|
| 1 | $H_2$/TH | 14 | $H_2$ | 27 | $C_2H_2$/THD |
| 2 | $CH_4$/TH | 15 | $CH_4$ | 28 | $H_2$/THH |
| 3 | $C_2H_6$/TH | 16 | $C_2H_6$ | 29 | $CH_4$/THH |
| 4 | $C_2H_4$/TH | 17 | $C_2H_4$ | 30 | $C_2H_6$/THH |
| 5 | $C_2H_2$/TH | 18 | $C_2H_2$ | 31 | $C_2H_4$/THH |
| 6 | $C_2H_2/H_2$ | 19 | TH | 32 | $C_2H_2$/THH |
| 7 | $C_2H_2/CH_4$ | 20 | THD | 33 | $H_2$/TCH |
| 8 | $C_2H_2/C_2H_6$ | 21 | THH | 34 | $CH_4$/TCH |
| 9 | $C_2H_2/C_2H_4$ | 22 | TCH | 35 | $C_2H_6$/TCH |
| 10 | $C_2H_4/H_2$ | 23 | $H_2$/THD | 36 | $C_2H_4$/TCH |
| 11 | $C_2H_4/CH_4$ | 24 | $CH_4$/THD | 37 | $C_2H_2$/TCH |
| 12 | $C_2H_4/C_2H_6$ | 25 | $C_2H_6$/THD | | |
| 13 | $(C_2H_4/H_2)+(C_2H_4/CH_4)$ | 26 | $C_2H_4$/THD | | |
| TH=$H_2+CH_4+C_2H_6+C_2H_4+C_2H_2$;THD=$CH_4+C_2H_4+C_2H_2$;THH=$H_2+C_2H_4+C_2H_2$; TCH= $CH_4+C_2H_6 C_2H_4+C_2H_2$ | | | | | |
| Above DGA Parameters ranked by skewness (lower to higher) | | | | | |
| | DGA parameters sorted by considering by higher to lower rank | | | | |
| Features No. from above | 28, 24, 1, 27, 31, 37, 26, 35, 36, 3, 32, 2, 34, 4, 5, 53,21,14,19,20, 13, 10, 23, 6, 22, 15, 18, 17, 7, 8, 16, 11, 9, 12, 25, 30 & 29 | | | | |

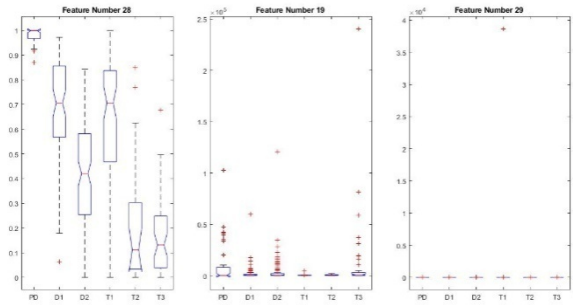

Fig. 1: Box plots of three DGA parameters (28, 19, 29 from Table-I) ranked by skewness

## 2.2 Intrinsic Time-scale Decomposition Based Feature Extraction and Optimal Feature Set Selection

In this Section, feature extraction by using intrinsic time-scale decomposition (ITD)-based feature extraction and optimal feature set selection are discussed.ITD is an efficient tool for extracting amplitude and frequency changing pattern from any nonlinear signal. In comparison, the well-known Empirical Mode Decomposition (EMD)has some shortcomings: (i) inaccurate outcomes when signal dynamics is considered, (ii) occasional failure to generate residual from IMF, when it has proper rotational property [8]. Thus, ITD is more effective than EMD for representing nonlinear and non-stationary signals and data decomposition [8-10].

Intrinsic time scale decomposition algorithm decomposes data series into integer sum of proper rotation components (PRCs) and residual signal. We decomposed each transformer's ranked features set into single-stage ITD domain and extracted the proper rotation component (PRC). For this purpose, initially we decompose the first eighteen ranked parameters of each transformer. Those eighteen DGA parameters are ranked considering lower to higher skewness and shown in Table II. We continue the process adding next DGA-ranked parameter and so on. Therefore, total nineteen PRC feature sets are obtained. There first set element is eighteen and last set element is thirty-seven. For getting optimal set of features, we classify those nineteen features set by XGBoost classifier among 376 transformer'sdataset, where training and testing are split in 85:15 ratio randomly. Those nineteen features set performance scenario is depicted in Fig.2.One can see that first ranked twenty-four extracted features provide the best classification performance. Thus, PRC coefficients of ranked twenty-four DGA features (i.e.28, 24, 1, 27, 31, 37, 26, 35, 36, 3, 32, 2, 34, 4, 5, 33, 21, 14, 19, 20, 13, 10, 23, 6) are our final feature set. The twenty-four features are then used in the classification schemes. Fig.3 shows that the PRC features provide good separation among classes. Moreover, the final twenty-four PRC features provide low p-values in general in one-way analysis of variance (ANOVA) test, average p-value being 0.0879.The thirty-seven DGA parameter is considered as thirty-seven data channels, each consisting of 376 data value. The ranking of the channels providesa pattern in which the statistically less informative channels are ordered after the more informative channels, thus providing better depiction of the fault classes. The low p-value of the features also indicates that. Note that ranking can also be done by Lasso, Laplacian score, Fisher score and ReliefF that have been used for optimal selection of DGA features [15].The average p-value for the twenty-four PRC features are found to be0.4789, 0.3267, 0.239 and 0.2479Lasso, Laplacian score, Fisher score and ReliefF, respectively, much higher than obtained with skewness. Thus, skewness-based ranking is more discriminative than the others.In the next Section, we discuss the classification procedure for fault diagnosis.

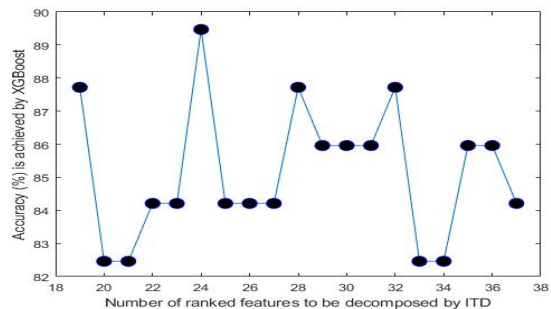

Fig. 2: Values of accuracy obtained by the XGBoost (for optimal PRC set selection)

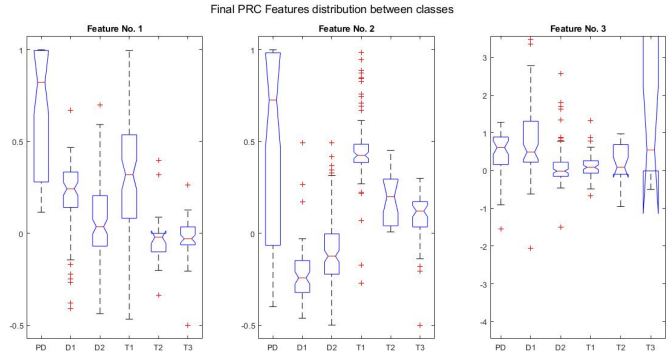

Fig. 3: Boxplots of the three PRC features (coefficients) among twenty-four among the six different classes of faults

### 2.3 Classification Method

For classification purpose, we use single XGBoost (Extreme Gradient Boosting) classifier which is widely used machine learning based classification schemes [7]. A single XGBoost classifier is used for detection of six class power transformer faults using the twenty-four PRC features described above. The implementation of XGBoost classifier is performed with default parameters in Python 3.7 environment. In the next Section, we will discuss classification performance in our proposed method with others conventional and machine learning based approaches.

## 3 Results and Discussion

In this Section, the experimental setting and performance analysis of the Proposed Method arepresented. The experiments are carried out in 4 GB RAM and 2.11 GHz Intel Core-i5 processor-based Windows-10 PC. This classification process is performed in Python 3.7 and feature extraction carried out in Matlab-2020b platform.

A total of 376 transformer DGA samples are used. Among them, randomly 333 is used for training and the rest 43 inthe test set.In Table-III, the confusion matrix is presented where in bracket the number of samples in each fault class is provided. The performance of the Proposed Method is investigated using accuracy, sensitivity and F1-score. The average sensitivity is 90% and the value of F1-score for each fault class is high, mostly near 1 or 1.

Table III: Confusion Matrix of Proposed Method (among random 43 samples)

| Predicted / Actual | PD | D1 | D2 | T1 | T2 | T3 | Sensitivity (%) | F1-score |
|---|---|---|---|---|---|---|---|---|
| PD (4) | 4 | 0 | 0 | 0 | 0 | 0 | 100 | 1 |
| D1 (7) | 0 | 7 | 0 | 0 | 0 | 0 | 100 | 0.9333 |
| D2 (15) | 0 | 1 | 14 | 0 | 0 | 0 | 93.33 | 0.9333 |
| T1 (6) | 0 | 0 | 0 | 6 | 0 | 0 | 100 | 1 |
| T2 (3) | 0 | 0 | 1 | 0 | 2 | 0 | 66.67 | 0.8 |
| T3 (8) | 0 | 0 | 0 | 0 | 0 | 8 | 100 | 1 |

The conventional method such as IEC method and Rogers four ratio method provides 'No Fault (NF)' and 'Undefined (UD)' states. Duval triangle makes false prediction in its boundary region. To illustrate this, the classification results of six randomly selected transformersare shown in Table IV. One can see that the Duval triangle can truly predict only one transformer fault class among six, Rogers four ratio method predicts 'Undefined' states on four samples while the IEC method predicts four 'UD' and two 'NF' falsely.On the other hand, the proposed method predicts the faults accurately.

Table IV: Proposed Method and Conventional Method Performance(Among Different Fault Samples)

| $H_2$ (ppm) | $CH_4$ (ppm) | $C_2H_6$ (ppm) | $C_2H_4$ (ppm) | $C_2H_2$ (ppm) | Actual Fault | Duval Method | Rogers four ratio Method | ICE Method | **Proposed Method** |
|---|---|---|---|---|---|---|---|---|---|
| 292 | 346 | 32 | 313 | 196 | D2 | D2 | UD | UD | D2 |
| 385 | 28.8 | 50 | 82.3 | 171 | D1 | D2 | UD | UD | D1 |
| 34 | 8.6 | 70.3 | 3.1 | 0.001 | T1 | T2 | T1 | NF | T1 |
| 157 | 46 | 76 | 12 | 0.001 | T1 | T2 | T1 | NF | T1 |
| 10 | 15 | 0.001 | 0.001 | 35 | D2 | D1 | UD | UD | D2 |
| 1651 | 90 | 33 | 45 | 2 | PD | T2 | UD | UD | PD |

Table V shows the performance comparison among various methods. In general the Proposed Technique give superior accuracy, about 32%, 9% and 4.65% higher than the conventional Duval Triangle Method, BA-PNN and EMD-based Hierarchical Ensemble Method, respectively.

Table V: Performance Comparison Among Different Fault Classes

| Fault Method | PD | D1 | D2 | T1 | T2 | T3 | Average Accuracy |
|---|---|---|---|---|---|---|---|
| Duval Method (%) | 25 | 42.86 | 73.33 | 66.67 | 33.33 | 87.50 | 62.79 |
| Rogers Four ratio method (%) | 0 | 0 | 66.67 | 100 | 33.33 | 25 | 44.19 |
| IEC method (%) | 0 | 28.57 | 53.33 | 66.67 | 33.33 | 62.50 | 46.51 |
| Ensemble Learning [7] (%) | 100 | 57.14 | 93.33 | 100 | 66.67 | 87.50 | 86.05 |
| BA- PNN [6] (%) | 75 | 42.86 | 66.67 | 66.67 | 66.67 | 62.5 | 62.79 |
| EMD based H-XGBoost[5] (%) | 100 | 85.71 | 93.33 | 83.33 | 100 | 87.50 | 90.70 |
| Proposed Method (%) | 100 | 100 | 93.33 | 100 | 66.67 | 100 | 95.35 |

The proposed Method is also more efficient computationally than the EMD-based technique. For training and testing, it requires 0.108 s and 0.00201 s, respectively, whereas the EMD-based one needs 0.456 s and 0.1456 s.Thelower time required and better accuracy for ITD-based method as compared the Hierarchical EMD-based technique can be attributed to; (i) the capacity of ITD of giving superior proper rotation components and hence, following the trend of a nonlinear data much better, and (ii) EMD employing spline interpolation requiring more memory and computation time. Moreover, the later Method uses three XGBoost classifiers in a hierarchical mode of classification.

As mentioned before, the Dataset used in this paper is imbalanced. This can bias the obtained accuracy towards the majority classes. F1-score and Cohen Kappa are well known matrices for performance evaluationin classification using imbalanced dataset[16]. Also, the Synthetic Minority Oversampling Technique (SMOTE) is invariably used with imbalanced dataset for experimentations in classification [17, 18]. A 5-fold cross validation is performed with SMOTE for the Proposed Method as well as the EMD-based Hierarchical Technique [5]. The cross-validation Techniques is employed to account for the relatively moderate size of the dataset. The Proposed Method yields anCohen Kappa and F1-Score of 0.91 and 0.92, respectively while the EMD-based Method gives 0.89 and 0.90.

## 4    Conclusion

A novel transformer fault diagnosis method has been proposed in this paper. The DGA parameters of publicly available 376 transformers have been generated and then ranked by their skewness. Intrinsic time-scale decomposition (ITD) has been used to extract features from skewness-based ranked parameters. The motivation for using ITD has been its improved capacity to capture the trend in nonlinear signals at reduced computational complexity as compared the well-known EMD. An XGBoost classifier has been employed to obtain the optimal set of extracted PRC features. The classification performance of the Proposed Method has been studied and compared with several existing DGA-based techniques. Conventional methodsgive 'No fault' or 'Undefined State' in many cases or suffers from modest accuracy. The Proposed Method overcomes these limitations and has provided more than 95% average accuracy as well as high F1-score and sensitivity in each fault classes, better than conventional methods and several machine-learning-based power transformer fault diagnosis method.Since the DGA dataset is slightly imbalanced, the effectiveness of the Proposed Method has been further investigated using a 5-fold cross-validation and SMOTE. The results have shown that the ITD-based proposed method providesa better performance than EMD-based hierarchical approach in terms of F1-score and Cohen Kappa score.